\documentclass[1p,11pt,times]{elsarticle}

\usepackage{ecrc}
\usepackage{epsfig}

\usepackage[cmex10]{amsmath}
\usepackage{amssymb}
\usepackage{amsthm}
\usepackage{algorithmic}
\usepackage{array}     	
\usepackage{subfigure}
\usepackage[normalem]{ulem}  
\usepackage[linesnumbered, ruled, vlined]{algorithm2e}
\usepackage{verbatim}  	
\usepackage{tabularx}  	
\usepackage{multirow}  	
\usepackage{booktabs} 	
\usepackage[table]{xcolor}  
\usepackage[bookmarks]{hyperref}
\usepackage{url}
\usepackage{epstopdf}

\jid{procs}

\usepackage{amssymb}

\usepackage[figuresright]{rotating}

\begin{document}

\begin{frontmatter}




\title{\bf Video Text Localization \\with an emphasis on Edge Features}


\author[rvt]{B.H.Shekar\corref{cor1}}
\author[els]{Smitha M.L.}

\cortext[cor1]{Corresponding author}

\address[rvt]{Department of Computer Science, Mangalore University, Mangalore, Karnataka, India. \\Email: bhshekar@gmail.com}
 
\address[els]{Department of Master of Computer Applications, KVG College of Engineering, Sullia, Karnataka, India.\\Email:smithaml.urubail@gmail.com}

\begin{abstract}
The text detection and localization plays a major role in video analysis and understanding. The scene text embedded in video consist of high-level semantics and hence contributes significantly to visual content analysis and retrieval. This paper proposes a novel method to robustly localize the texts in natural scene images and videos based on sobel edge emphasizing approach. The input image is preprocessed and edge emphasis is done to detect the text clusters. Further, a set of rules have been devised using morphological operators for false positive elimination and connected component analysis is performed to detect the text regions and  hence text localization is performed. The experimental results obtained on publicly available standard datasets  illustrate that the proposed method can detect and localize the texts of various sizes, fonts and colors.
\end{abstract}

\end{frontmatter}

\section{Introduction}
\label{section:Introduction}
The text present in images and videos contain lots of semantic information which are useful for video comprehension.
In recent years, the automatic detection of texts from images and videos has gained increasing attention. However, the large variations in text fonts, colors, styles, and sizes, as well as the low contrast between the text and the complicated background often make text detection extremely challenging. 

The existing methods of text detection and localization can be roughly categorized into two groups: region-based and connected component (CC)-based. Region-based methods attempt to detect and localize text regions by texture analysis. Generally, a feature vector extracted from each local region is fed into a classifier for estimating the likelihood of text. Then neighboring text regions are merged to generate the text blocks. These methods can detect and localize texts accurately  as the  text regions have distinct textural properties from the non-text even when the images are noisy. On the other hand, CC-based methods directly segment candidate text components by edge detection or color clustering. The non-text components are then pruned with heuristic rules or classifiers. Since the number of segmented candidate components is relatively small, CC-based methods have lower computation cost and the located text components can be directly used for recognition. 

The edge and gradient are the preferred features in text detection. Smith et al.\cite{Smith}, proposed  an  algorithm  to  detect  text  in individual video frames. Image blocks that including lots of sharp edges are considered as text. The algorithm is scale dependent, that is, only text with certain font size can be found. Sato et al.\cite{Sato}, proposed a method to detect edges that were connected to form text clusters by using smoothing filters. Cai et al.\cite{Cai}, proposed a text detection approach for video based on edge detection in YUV color space. Wu et al.\cite{Wu}, proposed an algorithm based on the image gradient produced by nine second-order  Gaussian  derivatives.  The  pixels  that  have  large gradient  are  considered  as  strokes.  These  strokes  are grouped into text blocks based on several empirical rules. Shivakumara et al.\cite{Shivakumara2010}, implemented a method based on Sobel edges and texture features for detecting text in video images.

Neha et al.\cite{Neha2012}, implemented a method where Sobel edge detector is applied on three detail components obtained after applying Discrete Wavelet Transform. The resultant edges so obtained were combined to form edge map which was further used for text localization. The texture-based methods\cite{Liu2005},\cite{hua2004automatic} assume text regions to have some kind of special textures. The texture-based methods are time-consuming and sometimes are influenced by the fonts and styles of characters. The stroke-based method\cite{epshtein2010} captures the intrinsic characteristics of text strokes so that the better detection results have been obtained even in complex background. CC-based approaches usually incorporate various methods such as edge detection, stroke width transform , or color clustering in order to localize the connected components. Some heuristic rules are employed to remove the false positives and some grouping techniques are imposed to form the text lines.
 
Although the existing methods have reported promising localization performance, certain problems need to be addressed. For region-based methods, the speed is relatively slow and the performance is sensitive to text alignment orientation.
On the other hand, CC-based methods cannot segment text components accurately without prior knowledge of text position
and scale. Moreover, designing fast and reliable connected component analyzer is difficult since there are many non-text
components which are easily confused with the texts when analyzed individually. Text localization in document/scene images and video frames aims at designing an advanced optical character recognition systems. The text localization and recognition in images and videos has received significant attention in the last decade\cite{Jain1998}. Hence, the video text localization and recognition is still an open problem. In this context, we propose a new approach for text localization  in videos and scene images. Our method aims to detect the text in the input image or the video frame by performing certain preprocessing. Further, the preprocessed image/video frame undergoes edge emphasizing, binarization, false positive elimination and text localization. The remaining part of the paper is organized as follows. The proposed approach is discussed in section 2. Experimental results and comparison with other approaches are presented in section 3 and conclusion is given in section 4.
\section{Proposed Methodology} 
\label{section:algo} 
\begin{figure}[hbtp]
\centering
\includegraphics[scale=0.4]{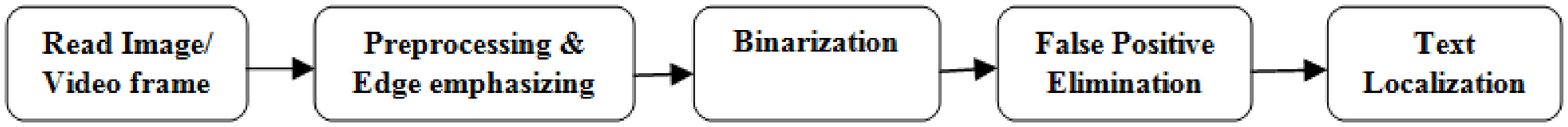}
\caption{Flowchart of the proposed system}
\label{fig1}
\end{figure}

 The flowchart of the proposed text detection and localization approach is shown in Fig.~\ref{fig1}. The details of each processing blocks are discussed below. 
\subsection{Preprocessing}  
\begin{figure}[hbtp]
\centering
\includegraphics[scale=0.4]{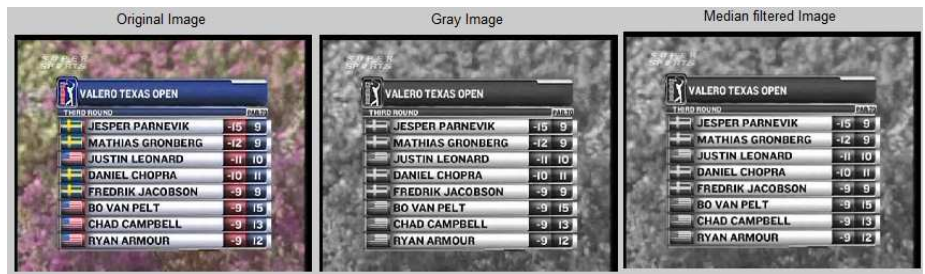}
\caption{Result of preprocessing}
\label{fig2}
\end{figure}

Text regions typically have a large number of discontinuities. In this phase, filtering is used to remove the unwanted noise in an image. The given input image is first converted into a gray image and median filtering is employed to remove the impulse noise in the resultant image. The main idea behind the median filtering is to blur the background which runs through the image pixel by pixel and to replace each pixel with the median of neighboring pixel. The characters are slightly blurred but the background areas are more blurred in the image as shown in Fig.~\ref{fig2}.
\subsection{Edge Emphasizing} 
\begin{figure}[hbtp]
\centering
\includegraphics[scale=0.4]{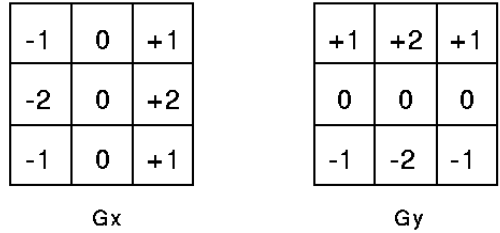}
\caption{Sobel Convolution Kernels}
\label{fig3}
\end{figure}

\begin{figure}[hbtp]
\centering
\includegraphics[scale=0.4]{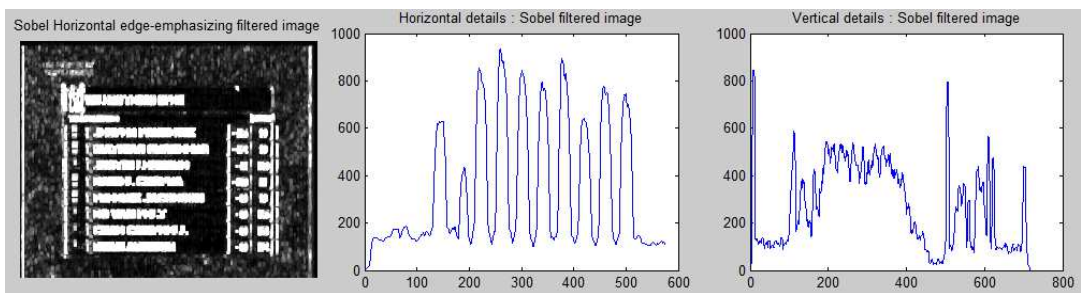}
\caption{Result of Sobel edge emphasizing}
\label{fig4}
\end{figure}

The edge is the distinct characteristic which is used to find the text features in an image. The edge of the text regions need to be detected to separate the background and the foreground. The edge of the image is computed using Sobel horizontal edge emphasis filter as shown in Fig.~\ref{fig3}. The  Sobel operator performs a 2-D spatial gradient measurement on an image and hence emphasizes regions of high spatial frequency that correspond to edges. Typically, it is used to find the approximate absolute gradient magnitude at each point in an input grayscale image.
The operator uses two 3 x 3 kernels as shown in Fig.~\ref{fig3} which are convolved with the median filtered image to calculate approximations of the derivatives in both vertical and horizontal directions. These kernels are designed to respond maximally to edges running in both of these directions. The kernels can be applied separately to the input image to produce separate measurements of the gradient component in each orientation called as Gx and Gy respectively. These can then be combined together to find the absolute magnitude of the gradient at each point. \\
  \begin{equation}
  \vert G \vert = \sqrt{Gx^{2} + Gy^{2}}  \\
  \end{equation}
Typically, an approximate magnitude is computed using:\\
\begin{equation}
\vert G \vert = \vert Gx \vert + \vert Gy \vert
\end{equation}
Since the Sobel kernels can be decomposed as the products of an averaging and a differentiation kernel, they compute the gradient with smoothing. The result of the Sobel operator is a 2-dimensional map of the gradient at each point. It can be processed and viewed as an image with the areas of high gradient are visible as text pixels. Now, we employ the sobel edge emphasizing filter to highlight the strong spatial gradients that correspond to edges as shown in Fig.~\ref{fig4}.
\subsection{Binarization}
We observe that the noise has increased during the edge detection and it is essential to remove all the noise pixels and at the same time retain the edge of the detected text. We now employ Otsu method of binarization\cite{Liu2009} to find the text clusters where we exhaustively search for the threshold that minimizes the intra-class variance. This method of binarization segments the text from the background as shown in Fig.~\ref{fig5}.
\subsection{False Positive Elimination}
The significant text regions obtained after binarization may also contain some non-text blocks. In order to improve the performance of the system these non text regions are removed using some geometric rules. First, we eliminate the pixels with intensity value less than 20 percent of the average area. The morphological dilation operation is performed to connect the broken components. We determine the connected components using 8-connectivity and compute the area of each component. The small objects are further removed using connected component labeling. It is observed that the text block should have more edge pixels than some non-text blocks. The foreground connected components in each of these frames are considered as text candidates as shown in Fig.~\ref{fig5}. 
\subsection{Text Localization}
\begin{figure}[hbtp]
\centering
\includegraphics[scale=0.4]{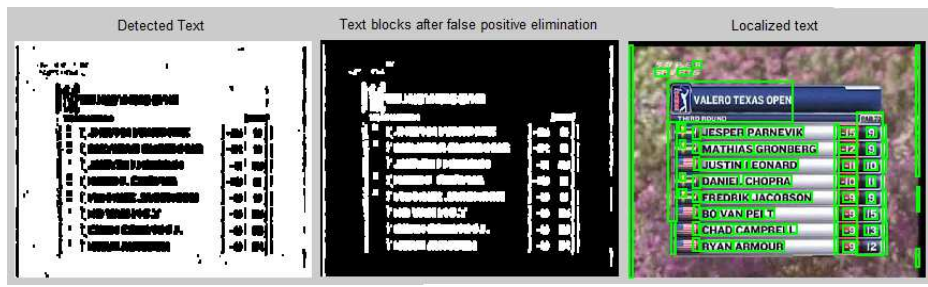}
\caption{Result of Text Localization}
\label{fig5}
\end{figure}

The objective of text localization is to place rectangles of varying sizes covering the text regions. We start by merging all the text detections. Further, the morphological dilation operation is performed to fill the gaps inside the obtained text regions which yields better results and the boundaries of text regions are identified as shown in Fig.~\ref{fig5}. Once the text regions are identified, the rectangles are placed to those text regions and finally localized.  
\section{Experimental Results}
\begin{figure}
\includegraphics[scale=0.3]{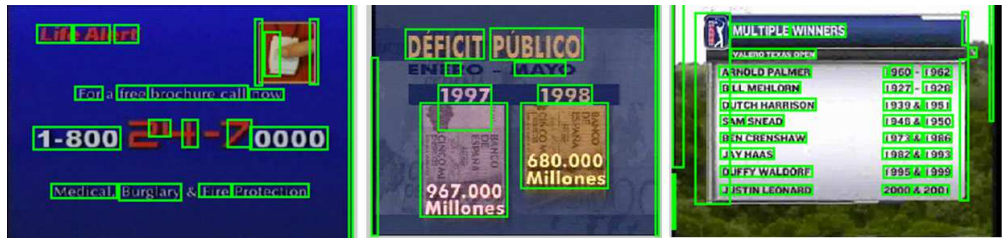}
\caption{Sample results of text localization}
\label{fig6}
\end{figure}

This section presents the experimental results to reveal the success of the proposed approach. The evaluation of the system on the various datasets show that it is capable of detecting and locating texts of different sizes, styles and types present in videos. The performance of the proposed approach is evaluated with respect to f-measure(F) which is a combination of two metrics:   precision(P) and recall(R).  The \textit{truly detected text block}($TDB$) is a detected block that contains partially or fully text. The \textit{falsely detected text block}($FDB$) is a block with false detections. The \textit{text block with missing data}($MDB$) is a detected text block that misses some characters. Based on the number of blocks in each of the categories mentioned above, the following metrics are calculated to evaluate the performance of the method. 
\begin{center}
 Detection rate = Number of TDB / Actual number of text blocks \\
 False positive rate = Number of FDB / Number of (TDB + FDB) \\
 Misdetection rate = Number of MDB / Number of TDB
\end{center}
We have conducted experiments on ICDAR text locating competition dataset\cite{Lucas2005} and video datasets such as Hua, Horizontal-1 and Horizontal-2 which are said to be the bench mark datasets and the results are reported in Table~\ref{tab1} and Table~\ref{tab2}.  Sample results of the localized text blocks  are shown in Fig.~\ref{fig6}. In order to exhibit the performance of the proposed approach, we have also made a comparative analysis with some of the well known algorithms\cite{Liu2005},\cite{Wong03},\cite{Trung},\cite{Kasturi} which is on-par with the state-of-the-art text localization approaches.

Liu et al.\cite{Liu2005}, devised an edge-based method that extracts edge features by using the sobel operator. Wong et al.\cite{Wong03}, proposed gradient based  method that  computes  the  MGD  values  to  identify candidate text regions. Trung et al.\cite{Trung}, proposed laplacian based approach that used MGD values to detect the text clusters. Mariano et al.\cite{Kasturi}, proposed uniform-colored method that performed clustering in the L*a*b* color space to locate the text lines. 

\begin{figure}[hbtp]
\centering
\includegraphics[scale=0.4]{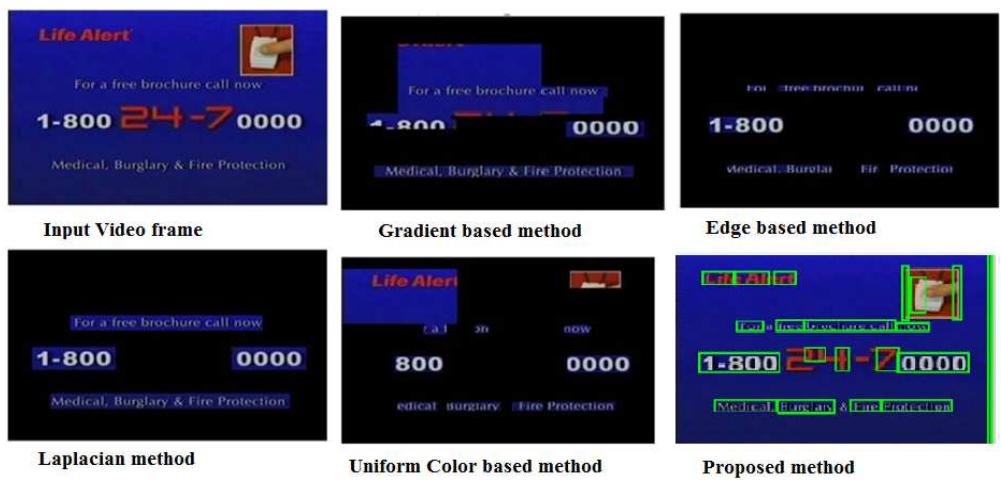}
\caption{Results of text localization of various approaches}
\label{fig7}
\end{figure}

Fig.~\ref{fig7} shows some sample results of the existing methods and the proposed method for comparative analysis. The edge-based method fails to detect some text blocks because of the problem of fixing threshold values for edge detection. The gradient-based method detects the text blocks with missing characters and inaccurate boundary. This method uses many threshold values and heuristic rules and thus may only work well for specific datasets. The  uniform-colored  method  detects  the  text blocks with missing characters and produces many false  positives  due  to  the  problem  of  color
bleeding.  The  laplacian  method  detects  all  the blocks correctly and even picks up the low contrast logo of the television. Our proposed approach has successfully detected all the text but has some false positives as shown in Fig.~\ref{fig7}.  The proposed approach has gained slightly improved precision and recall rates on Horizontal-1 video dataset and comparable results for other datasets when compared with other works in literature. 
\begin{table}
\centering
\caption{Evaluation performance of the proposed method on various datasets }
\begin{tabular}{|c|c|c|c|c|}
\hline 
 \cline{2-4}
Dataset & Recall & Precision & F-measure  \\ 
\hline 
 Hua & 0.7845 & 0.6848 & 0.7300  \\ 
\hline 
Horizontal-1 & 0.8734 & 0.7423 & 0.8025  \\ 
\hline 
Horizontal-2 & 0.6489 & 0.7849 & 0.7105  \\ 
\hline 
ICDAR 2011 & 0.7934 & 0.7245 &  0.7574 \\
\hline
\end{tabular} 
\label{tab1}
\end{table}
\begin{table}
\centering
\caption{Evaluation performance of various approaches for Horizontal-1 dataset }
\begin{tabular}{|c|c|c|c|c|}
\hline 
 \cline{2-3}
Methods & R & P   \\ 
\hline 
Edge based\cite{Liu2005}   & 0.80 & 0.18   \\ 
\hline 
Gradient based\cite{Wong03} & 0.71 & 0.12   \\ 
\hline 
Uniform based\cite{Kasturi} & 0.51 & 0.27  \\ 
\hline 
Laplacian based\cite{Trung} & 0.83 & 0.71   \\ 
\hline 
Proposed method & 0.87 & 0.74   \\ 
\hline
\end{tabular} 
\label{tab2}
\end{table}
\section{Conclusion}
\label{section:conclusion}
Text in video is useful and important in indexing and retrieving the video documents efficiently and accurately. We developed a sobel edge emphasizing based approach that detects the text clusters. The newly developed approach is capable of localizing the text regions in  video frames.  Experimental results show that the proposed method is effective for identifying text blocks. The robustness of the proposed method against noise and low contrast of
text is demonstrated by false positives in the dataset. The various problems that occur due to complex background need to be addressed in our future works.

\end{document}